# Taxonomy and evolution predicting using deep learning in images

Jiewen Xiao[1, #], Wenbin Liao[2, #], Ming Zhang[4], Jing Wang[3], Jianxin Wang[1, 2, *], Yihua Yang[3, *]

[1] College of Environmental Science and Engineering, Beijing Forestry University, 35 Qinghua East Road, Haidian District, Beijing 100083, P. R. China

[2] School of information Science and Technology, Beijing Forestry University, 35 Qinghua East Road, Haidian District, Beijing 100083, P. R. China

[3] Guizhou Institute of Biology, Guiyang National Economic and Technological Development Zone, Guizhou Province 550000, P. R. China

[4] State Key Laboratory of Applied Microbiology Southern China, Guangdong Provincial Key Laboratory of Microbial Culture Collection and Application, Institute of Microbiology, Guangdong Academy of Sciences, Guangzhou 510070, China

#: the contribution of authors is equal.

*Corresponding author:

Yihua Yang

E-mail: yangyh_009@163.com

Jianxin Wang

E-mail: wangjx@bjfu.edu.cn

**Abstract**

Molecular and morphological characters, as important parts of biological taxonomy, are contradictory but need to be integrated. Organism's image recognition and bioinformatics are emerging and hot problems nowadays but with a gap between them. In this work, a multi-branching recognition framework mediated by genetic information bridges this barrier, which establishes the link between macro-morphology and micro-molecular information of mushrooms. The novel multi-perspective structure is proposed to fuse the feature images from three branching models, which significantly improves the accuracy of recognition by about 10% and up to more than 90%. Further, genetic information is implemented to the mushroom image recognition task by using genetic distance embeddings as the representation space for predicting image distance and species identification. Semantic overfitting of traditional classification tasks and the granularity of fine-grained image recognition are also discussed in depth for the first time. The generalizability of the model was investigated in fine-grained scenarios using zero-shot learning tasks, which could predict the taxonomic and evolutionary information of unseen samples. We presented the first method to map images to DNA, namely used an encoder mapping image to genetic distances, and then decoded DNA through a pre-trained decoder, where the total test accuracy on 37 species for DNA prediction is 87.45%. This study creates a novel recognition framework by systematically studying the mushroom image recognition problem, bridging the gap between macroscopic biological information and microscopic molecular information, which will provide a new reference for intelligent biometrics in the future.

## Main

Organism's image recognition, including microbes[1], insect[2], butterfly,[3] plant[4,5], birds[6], animal[7], and *et al.*, often as a fine-grained image recognition task[8], have achieve great progresses with aids of machine learning (ML) and deep learning (DL) algorithms. Organism-specific image identification will contribute to the automation of ecologic and taxonomic research. In our work to integrate image recognition and genetic information prediction of organisms, macro-fungi, namely mushrooms, are used as the object of study, for they are widely distributed[9], have important ecological functions, especially interact with microorganisms and plants[10,11], are diverse in species, are easy to collect and image, and have not been extensively studied. Furthermore, edible mushrooms as a great source of food for humans, but are often confused with highly toxic mushrooms causing serious organ damage and even death.

## Multi-perspective recognition framework

Since general single-branch model only focus on one region of the sample, it greatly limits its ability to extract target features. Thus, multi-branch models[12-14] are of great help to the model performance improvement. However, restricted to the composition of the dataset from a single perspective, multi-branching models also tend not to perform optimally. We have specifically designed a dataset with multi-perspective image, then we fuse the features captured by the three branches and use a classifier to make the final recognition of the mushrooms (Fig. 1).

We take images of the mushroom from 3 perspectives of top view (pileus), side view (lamella and stipe), and bottom view (lamella and stipe) from different angles in 3 axises of two sides (Fig. 1a). *Rugiboletus extremiorientalis* is shown as an example sample in Fig.1b. Due to the small amount of data in our dataset, in order to make the model training better, we firstly used the migration learning method to load the trained model parameters from ImageNet to our branching model. As each perspective of the mushroom image corresponds to a branch, we constructed a total of three branches, froze most of the convolutional layers of the branching model, and only kept the last convolutional layer and the classification layer for fine tuning (Fig. 1c). In fact, this step of fine-tuning can be considered as the training of general single-branch models. After all, three branch models are trained, we discard the classification layer and fuse the features of the three branches using concat, and then classify the fused features by a classifier at the end (Fig. 1d). In the feature fused step, we freeze all the parameters of the branches and only fine-tune the classifier, which consists of an attention model (a SE module and an ECANet module) and a fully connected (fc) layer followed by an activation function.

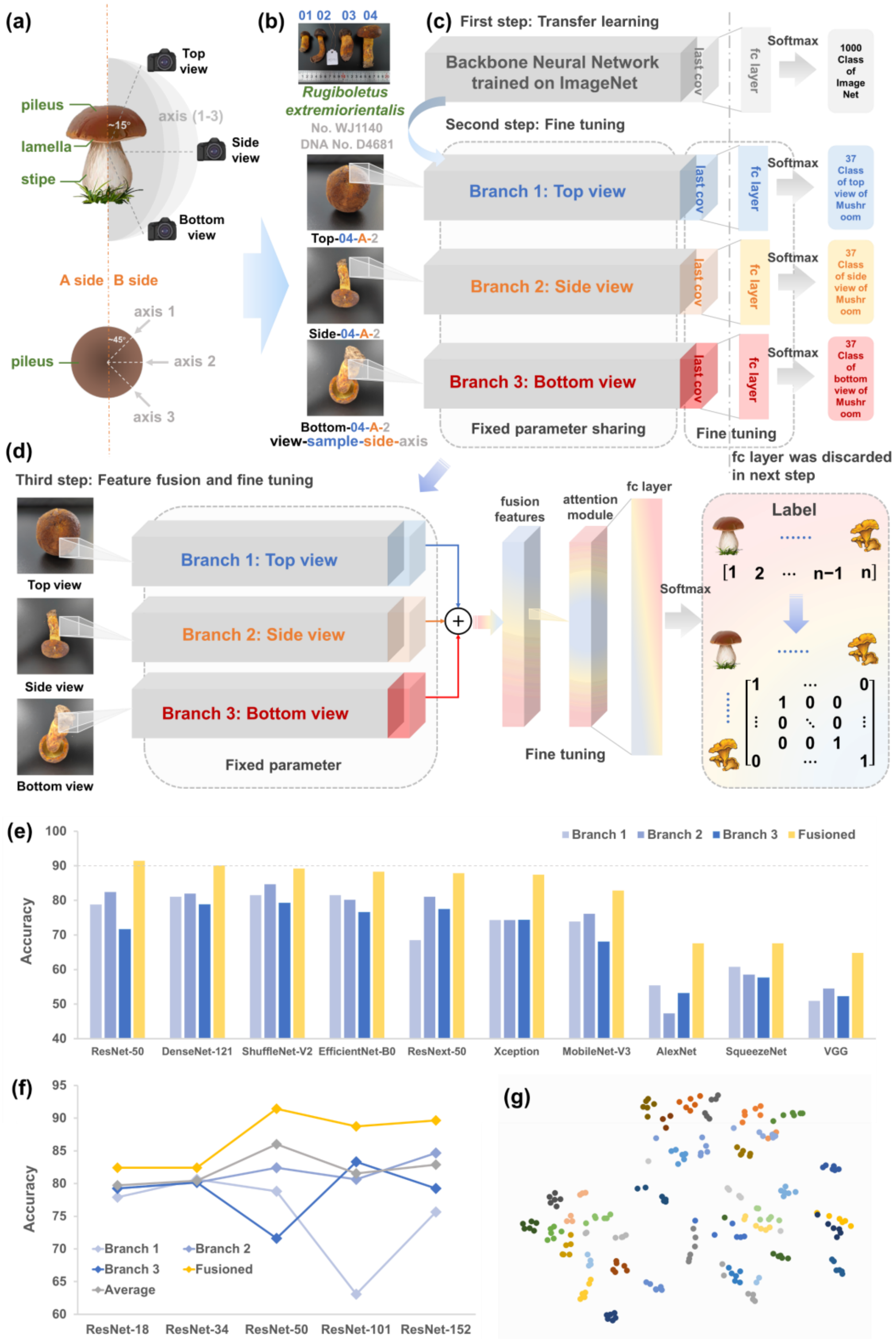

**Fig. 1 Schematic of a multi-perspective framework for mushroom recognition.**

(a) Multi-view mushroom image acquisition method. (b) Example of a multi-view mushroom image (*Rugiboletus extremiorientalis*). The transfer learning strategy (c) on our multi-perspective framework and multi-view mushroom image recognition framework fine-tuning (d). (e) The recognition accuracy of different backbones loaded onto multi-target recognition framework on the test set. (f) The test accuracy of ResNet loaded onto a multi-target recognition framework with different layers on the test set. (g) The t-sne diagram of the output of the ResNet-50 model.

To explore the effect of different backbone models, all 37 species of this project were formally experimented, and we selected VGG, SqueezeNet, Exception, DenseNet 121, Resnet50, ResNext50, ShuffleNet-V2, MobileNet-V2, MobileNet-V3 and EfficientNet-B0, and other models as backbone. The accuracies of single-branch models tested in the second step of fine-tuning exhibit more average performance in three perspectives of mushroom data on one backbone (Fig. 1e). Among them, the average accuracy of side view on 10 backbones is 72.11% relatively high, and the average accuracy of bottom view is 68.91% relatively low. But it is shown that with only a single input, the mushroom image features obtained from different views have no significant ($p$ value > 0.1) impact or difference for the recognition task (Supplementary Table 10). We found that ResNeXt-50 with the attention module did not perform better than ResNet-50. Among the tested models, ShuffleNet-V2 has better performance, the accuracy on the three branches is 81.53%, 84.68%, and 79.27%, correspondingly. The accuracy rate increased significantly ($p$ value < 0.1) after branch fusion (Supplementary Table 11), indicating that the multi-perspective structure has better performance than the single-perspective model. Among them, the accuracy rate of ResNet-50 was 91.89% in the validation set and 91.44% in the test set after fusion, which reached the best level among the tested models. After branch fusion, the classifier fused information from multiple views to make more accurate classification. The differences between the models are relatively significant ($p$ value < 0.05), with both the advanced deep model and the lightweight model working better (Supplementary Table 12). Moreover, it can be seen that during the training process, the increase in accuracy and decrease in loss of the three branches (ResNet-50 for example) fluctuate more and converge more slowly (Supplementary Fig. 1). But the fused fine-tuned model fluctuates less and converges faster, which is significantly better than the single-branch models.

For the branch model, the performance under multi-perspective structure is different, in order to backbone model the impact and effect of different structures, we selected ResNet as the backbone, set 18 layers, 34 layers, 50 layers, 101 layers, 152 layers for the study, the performance of different layers model is not very different at 0.1 level of p (Fig 1f, Supplementary Table 13) in 3 branches but significant in fine-tuning fusion model or average ensemble model in second step at 0.05 level. In the tested fine-tuning fusion model, the model with 50 layers has the highest accuracy, indicating that too many or too few layers can not improve the feature extraction ability of the branch model. The accuracy of fusion model is significantly better ($p<0.05$) than single-branch models and average ensemble model integrated by branching models directly (in Supplementary Table 14). The t-sne clustering results for the output of the model showed a more dispersed distribution among species (Fig. 1g).

**Image recognition meditated by genetic information**

Traditional taxonomy relies on morphological dichotomies to classify species, a very difficult task that requires taxonomists with a large training and knowledge base. The aforementioned fine-grained image recognition approach based on a multi-perspective framework can help automate taxonomy tasks. Recent modern taxonomy has introduced more genetic and genomic based phylogenetic tools. Where morphology and molecular-based phylogenies conflict, classification often prevails over more advanced genetic and genomic information. Even the identification of some species now relies solely on information obtained from molecular biology.

Deep learning[15-17] has been widely used in the study of biologically relevant genes and genomes for protein structure prediction[18,19], protein function prediction[20,21], genetic engineering[22,23], etc[24]. But in the problem of pattern recognition, there are few deep learning approaches that use genetic and genetic information to study. We note that genetic distances can be easily obtained from gene and genome calculations and contain evolutionary and taxonomic information, which is suitable for biologic pattern recognition. Therefore, in this work, we add genetic distance to the image pattern recognition task, breaking down the barrier between morphology and molecular phylogenetics.

What's more, traditional image recognition is difficult to measure the inter-image distance, and the image distance calculated by a given algorithm is difficult to have the correct meaning in real scenarios, especially in the field of biometric recognition, where it is difficult to measure the difference between images in accordance with the biological semantics. But genetic distance is uniquely positioned to solve this problem.

We use genetic distance sapce as a representation space and uses the genetic distances between pairs of mushroom species within the dataset to form embeddings of the genetic distance representation space. Genetic distance embeddings could be predicted through CNNs models based on images for image distance calculation and species prediction. In this study, image features are extracted using convolutional neural networks, which map image features to the feature space of the fully connected layer (fc). Then the feature space connects with the representation space (genetic distance embedding generated from the ITS region of DNA) through loss functions of softmax or MSE function (Fig. 2a). By learning on the training set, the model gains the ability to generate genetic distance through exacted image features, namely image distance with biogenetic information is obtained through image feature extraction. The category label embedding is generated by calculating the distance between the predicted genetic distance embedding and the true genetic distance embedding. And the true category corresponding to the image can be predicted by mapping the category label embedding to the category (Fig. 2b).

The 46 accurately identified subsets of dataset are used for the study, of which 9 subsets have duplicate species with other subsets. When using the 46 × 46 genetic distance matrices before censoring as the genetic distance space for 46 categories, the test accuracy was 55.79% (Supplementary Table 15), and we found that duplicate species in the training set would lead to imbalance in the training set. So, there are 37 species after reducing the duplicate subsets. However, the genetic distance can be calculated using duplicate species pairs, so the genetic distance matrix for 37 species is 37 × 46. The results of this study using 37 different species in the dataset are shown in Supplementary Table 15 and Supplementary Fig.2, and it can be found that the imbalance caused by the redundant training sets leads to slightly lower recognition accuracy, but the genetic distance recognition error of different subsets of the same species is not significant, and it can still correctly recognize species classified as different classes at training time. However, under the traditional CNN recognition method, if different subsets of the same species are classified as different, it leads to serious misclassification. The next study reveals the overfitting in the category space and underfitting to the semantics caused by softmax and one-hot labels in the traditional CNN model.

Using the MSE function, the prediction of genetic distance is very good for both averaging (MSE-mean) with $R^2$ of 0.867 and summation (MSE-sum) with $R^2$ of 0.893 (Supplementary Table 16 and Supplementary Fig. 3). The accuracy (root mean square error, RMSE) of predicted genetic distance is 0.0364 (95% confidence interval = 0.0345–0.0383) by using MSE-sum (Fig. 2d) and the accuracy for MSE-mean is 0.0343 (95% confidence interval = 0.0326–0.0360) whereas the accuracy (error of 100 replications by Bootstrap method) of the genetic distance calculated by the maximum composite likelihood method is 0.0349 (95% confidence interval = 0.0346–0.0353). Therefore, we made accurate predictions of the genetic distance of mushroom images at an accuracy close to that of genetic distance generation. However, compared with the model using softmax function, MSE is slightly less accurate in species identification (Fig. 2g). Using Softmax is able to make better species prediction with accuracy of 72.97% (Fig. 2g), but we find that when it leads at the expense of the accuracy of the genetic distance test, namely $R^2$ is 0.479 and RMSE is 0.3549 (95% confidence interval = 0.3260–0.3838) (Supplementary Table 16), huge errors near the diagonal (genetic distance is 0) can be observed in Fig. 2e and Supplementary Fig. 4. Among them, the diagonal element errors are the largest from -1 to -7, which is the opposite of generating genetic distances where the greater the genetic distance the greater the error. The zero element of the diagonal of the genetic distance matrix represents the distance between the species themselves and themselves. Therefore, we believe that the large errors around the diagonal of the genetic distance matrix represent an overestimation of the true distance between species by the model, which does improve the accuracy of the classification, but has lost much performance in the estimation of the semantics. In particular, when the diagonal element is set to -1, namely softmax(-1), the softmax function is trained to zoom in on the distance between images, overfitting to the label space by sacrificing the estimation of image distances in a higher dimensional representation space, resulting in a very high accuracy of about 90% (Fig. 2g), close to the case where no genetic distance representation space is added. However, it completely lost the ability to fit the genetic distance with an $R^2$ of 0.194 and an error of up to 3.767 (95% confidence interval = 3.686–3.847). From Fig. 2e as a whole, the cost of linearly improving recognition accuracy is an exponential increase in genetic distance prediction error. From Figs. 2h to 2j, the points in the dashed boxes are the predicted genetic distance embedding of the Russulaceae (including *Russula* spp., *Lactatius* spp. and *Lactifluus* spp.) after t-sne dimensionality reduction, and the results show the expansion of this taxon in the feature space. But surprisingly, even though the softmax(-1) model produces a huge expansion compared to the MSE model, the aggregated topology is still maintained with this Russulaceae, which exists dispersed in the prediction of the model without the use of genetic distance. Therefore, in most image classification tasks, models are overfitted in the label space to achieve accurate identification on species within the training set, which is detrimental to both knowledge learning and generalizability of the model. From the box plot of the error of the MSE (Figure 2k), the precision is significantly (at the 0.05 level) different among the species, and found larger errors for species with overall greater genetic distance (more distant phylogenetic relationships) (Supplementary Fig. 3).

In this study, four different distances were used to calculate the distance between the genetic distance of the predicted images and the true Ground Truth, with the best results using the cosine distance of 73.87%, whose ROC plots are shown in Supplementary Fig. 4. It shows that different distance measures have an effect on the space of label of the representation space mapping of genetic distance.

This study used genetic distances generated from DNA from public datasets with an accuracy of 75.22% (Supplementary Table 17), which is good, so this model can be experimented with DNA data extracted from non-mushroom samples, even slightly higher than using locally extracted DNA data. The premise of using public DNA data to construct genetic distances is that the species identification needs to be very accurate for the training set.

Based on the above findings, this study constructs different lengths of embeddings by using publicly available DNA sequences outside the dataset. Therefore, this study can further investigate the effect of different embedding lengths on the experimental accuracy and find that a longer embedding can improve the accuracy of the model (Fig. 2l). When the length of embedding is 100, which represents the genetic distance between the mushroom species corresponding to the changed mushroom image and 100 mushroom species, is based on the increase of local 46 subsets to 100 species, the accuracy of its classification task test can reach 70.72%, which is about 5% higher than the genetic distance between local species with the original length of 46, and when the Agaricomycetes of 3000 mushroom species DNA was used to construct the genetic distance dataset, a test set accuracy of 81.53% could be achieved. When the embedding length was significantly increased, there was no significant decrease for the overall predictive power of the model for genetic distances. Further, this study investigated the effect of shorter embedding length on the experimental accuracy and found that shorter embedding significantly reduced the accuracy of the model. From the t-sne results, too short genetic distance embedding cannot distinguish species from each other, while longer genetic distance can increase the feature space and improve the classification ability of the model. And it is easy to increase the genetic distance embedding length, which can use species sequences outside the dataset and improve the model's ability to learn more advanced semantics. However, in this study, using Gblocks to automatically crop DNA sequences, long genetic distances need to use more DNA for calculation, and using too many DNA sequences will lead to shorter retained DNA sequences and weaken the semantic information of genetics and evolution, so a more reasonable sequence crop strategy should be considered in the case of multiple sequences.

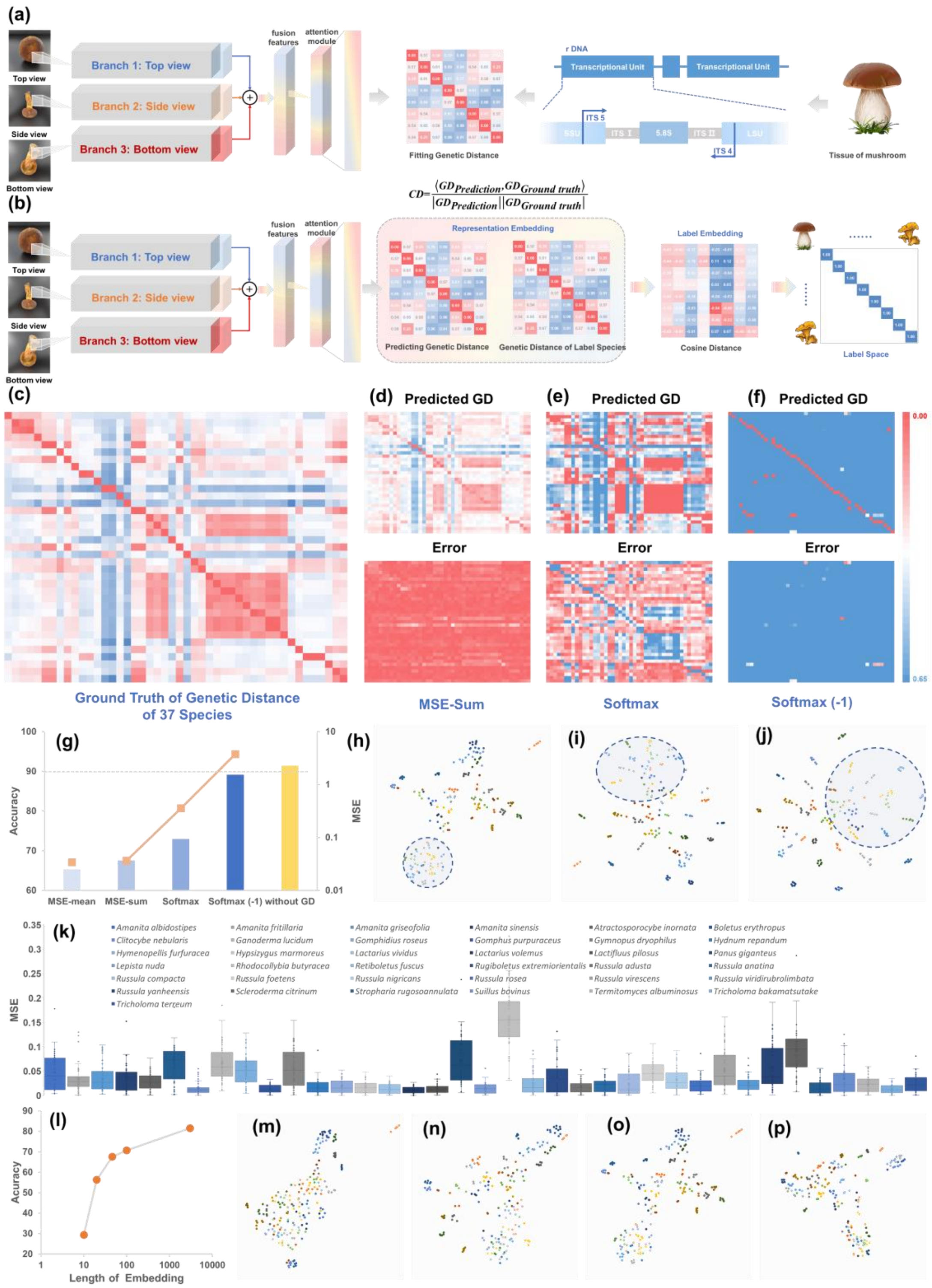

**Fig. 2 Introduction of genetic distance for mushroom image recognition.**

(a) Learning to generate genetic distance matrices from ITS sequences of DNA using a multi-perspective framework. (b) Predicting mushroom species through the medium of genetic distance matrix using a multi-perspective framework. (c) Ground truth of genetic distance of 37 species. Predicted genetic distance matrix and errors by ResNet-50 backbone inducing MSE-sum (d), Softmax (e), and Softmax(-1) (f) active function. (g) Test accuracy of mushroom species prediction and MSE of genetic distance prediction using ResNet-50 backbone with different activation functions (blue), and test accuracy of mushroom species prediction using ResNet-50 backbone without genetic distance (yellow). The t-sne diagrams of genetic distance embedding prediction using ResNet-50 backbone inducing MSE-sum (h), Softmax (i), and Softmax(-1) (j) active function. (k) Box line diagram of genetic distance prediction errors using ResNet-50 backbone inducing MSE-sum active function. (l) Test accuracy of mushroom species prediction by using genetic distance embeddings with different lengths from10 ~ 3000. The t-sne diagrams of genetic distance embedding with different lengths, namely the length of 10 (m), 20 (n), 43 (o), and 100 (p).

## Granularity issues for fine-grained image recognition

The division of different levels in taxonomy may have an impact on the accuracy of the classification task, especially for fine-grained classification problems, where the division of classification levels is the granularity level of fine-grained classification. In this study, samples at the genus level of hierarchy are selected from a multi-view mushroom image dataset for the classification task, and a total of 27 genus images are used for the experiments. The test accuracy of 92.36% indicates that species prediction can be little better achieved at the genus level with the multi-perspective structure-based model than mixing level (inter- and intra-generic mixing) with accuracy of 91.44% (Fig. 3a). For the model that used genetic distance, there was also some improvement in genus-level classification performance compared to the mixed classification, with accuracy increasing from 66.67% to 72.92% and AUC increasing from 0.93 to 0.95. As can be seen by Figs. 3c-e, the closer to the phylogenetic root the intergeneric identification is, the greater the error. As seen in Figure 3f, confusion also exists for intergeneric identifications that are genetically close to each other. As can be seen by Fig. 3g, Russulaceae genera remain clustered in the genetic distance-mediated model with Amanita spp. in the middle, while in Fig. 3h, Russulaceae genera have dispersed and *Amanita* spp. are at the edge.

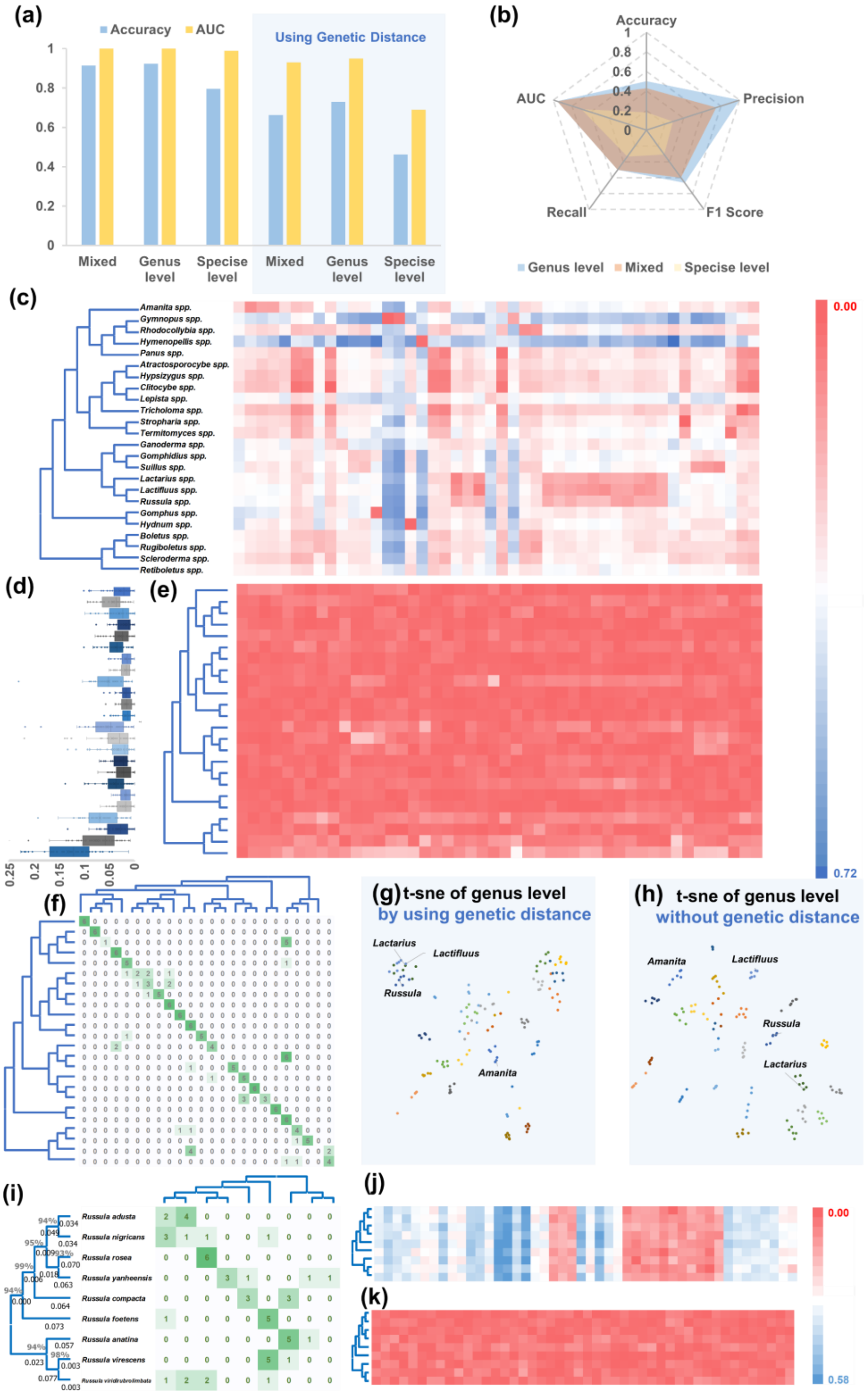

**Fig. 3 Fine-grained image recognition at species and genus level.**

(a) The test accuracy and AUC of fine-grade mushroom image recognition at mixed, species (*Rusulla spp.*), and genus level, using ResNet-50 backbone (left), and ResNet-50 backbone inducing genetic distance (right). (b) The rose chart of test accuracy, AUC, precision, recall, and F1 score for mushroom image recognition using ResNet-50 backbone genetic distance at mixed, species (*Rusulla spp.*), and genus level. The predicted genetic distance (c), errors (e), box line diagram of errors (d), and confusion matrix (f) using ResNet-50 backbone genetic distance at genus level with phylogenetic trees. The t-sne diagram of genetic distance embedding (g) and output of models without genetic distance (h) at the genus level. The confused matrix (i), predicted genetic distance (j), and errors (k) using ResNet-50 backbone genetic distance at species (*Rusulla spp.*) with phylogenetic trees.

Species within the same genus are more difficult to classify due to the problem of similarity, so nine species under the genus *Russula* spp. were selected for an interspecific classification study within the genus in this study. As shown, the accuracy of the interspecific classification task under the genus *Russula* spp. decreased to 79.62%, indicating that there is some error in the interspecific classification within the same genus. In the genetic distance-mediated model, the recognition within the genus was even worse (Fig. 3i and Supplementary Table 18), with an accuracy of only 46.30% and an AUC as low as 0.69, while the prediction accuracy for genetic distance is still high.

We can explain from this perspective that studying mushrooms from their macroscopic morphology is generally only possible to determine to the genus level. Because the genetic distances of species within the genus are relatively small (<0.1), close to the accuracy of predicting genetic distances from images. In particular, with a genetic distance of 0.003 between *R. virescens* and *R. viridirubrolimbata*, it is not possible to identify two species using image embedding, and the classification accuracy of these two species is 0. In the model without genetic distance, the effective identification of species within the genus is due to the overfitting of the labels, which is not a good performance in terms of interpretability. In the future, to accurately identify interspecific species within a genus, it is necessary to introduce microscopic images to obtain more subtle features, such as spore morphology. And the error in generating genetic distances by the maximum composite likelihood method is linearly related to genetic distances, so phylogenetic methods based on genes and genomic sequences can achieve high accuracy between species that are very close to each other to ensure accurate identification at the species level. Based on our results, we can easily "barcode" organisms at the genus level by quickly predicting the genetic distance embedding in biological images.

**Generalization and zero-shot identification**

In the zero-shot recognition task, auxiliary information is used to better help the model learn features, including attributes, text, word vectors, and recent DNA barcodes[25,26]. Badirli et al.[25,26] used the parameters of the fc layer of the DNA classifier or the parameters of the mixed fc layer of DNA classifier and image classifier as embedding, which achieved good results but were still poor in interpretability and their use of DNA barcodes lacked sufficient biological significance.

In this study, we firstly investigated the effects of adding two untrained species (generating genetic distances also avoids their DNA), *Thelephora aurantiotincta* (Fig. 4b), and *Amanita vaginata* (Supplementary Fig. 5), to the test set on the experiments, and found that the effects on the model accuracy were small, but the addition of *Amanita vagunata* significantly reduced the accuracy of the model (Supplementary Table 19). Among them, the AUC of *Theleora aurantiotincta* can reach 0.97 (Fig. 4c-d), which indicates that the model has some ability of zero-sample learning and can be further experimented. We perform a t-sne dimensionality reduction of predicted genetic distance embeddings and real genetic distance embedding of *Theleora aurantiotincta* along with the genetic distance embeddings of all seen species, and find that *Theleora aurantiotincta* is relatively close to its label (Fig. 4e). However, the identification of *Amanita vagunata* is worse, and it is almost impossible to identify it (Supplementary Table 20). From the t-sne, the distribution of *Amanita vagunata* is more scattered, partly near *Amanita* spp. and partly at the edge, which is far from the real genetic distance (Supplementary Fig. 6). The prediction accuracy of genetic distance for *Theleora aurantiotincta* is higher at 0.0418 (Fig. 4f) and slightly lower at 0.0772 for *Amanita vagunata* (Supplementary Fig. 7). We constructed phylogenetic trees using the predicted genetic distance (Fig. 4g) and the true genetic distance (Fig. 4f) of *Theleora aurantiotincta* with the genetic matrices of the other 43 species, and the only difference between the predicted and true phylogenetic trees is that the predicted branch length of *Theleora aurantiotincta* is shorter by 0.04, and this error matched the accuracy of the genetic distance. The main reason for the large identification error of the zero sample of *Theleora aurantiotincta* and *Amanita vaginata* is that *Amanita vaginata* is a species belonging to *Amanita* spp. already present in the dataset.

Then, five untrained species are used as the zero-sample training set, and the other 32 species are used as the training set, and the experimental results are shown in Fig 4 and Supplementary Table 20, where zero-shot learning can achieve an accuracy rate of 39.58%. Among them, the highest AUC of *Lactifuus pilosus* was 1.00. We visualize the predictions for the zero-shot task, Figure 4i shows that all five classes are well classified, and using the k-mean method we were able to obtain 98.33% accuracy for all five classes. However, what we would like is for these unseen classes to be accurately mapped to their true genetic distance neighborhood. Therefore, we added the seen samples and true genetic distances to visualize the unseen samples (Fig. 4j). Therefore, we added seen samples and true genetic distances to visualize the unseen samples. We found that *Lactifuus pilosus* is close to the true genetic distance, while it other species were far from the true sample.

In this study, based on the aforementioned genetic distances predicted by image features, a wide range of features of mushroom species can be learned, and features can be mapped to a genetic distance representation space containing other species besides the training species, thus allowing zero-shot learning. And, our method does not require the use of surrogate methods to make predictions for unseen samples. The Embedding generated by image feature space mapping can be compared with the genetic distances of species within the non-training set to calculate cosine distances, Euclidean distances, etc., so that an unseen label embedding can be generated directly and mapped to the zero-shot label space. And we recommend the use of predicted genetic distances to directly construct phylogenetic trees for initial species identification rather than direct species identification.

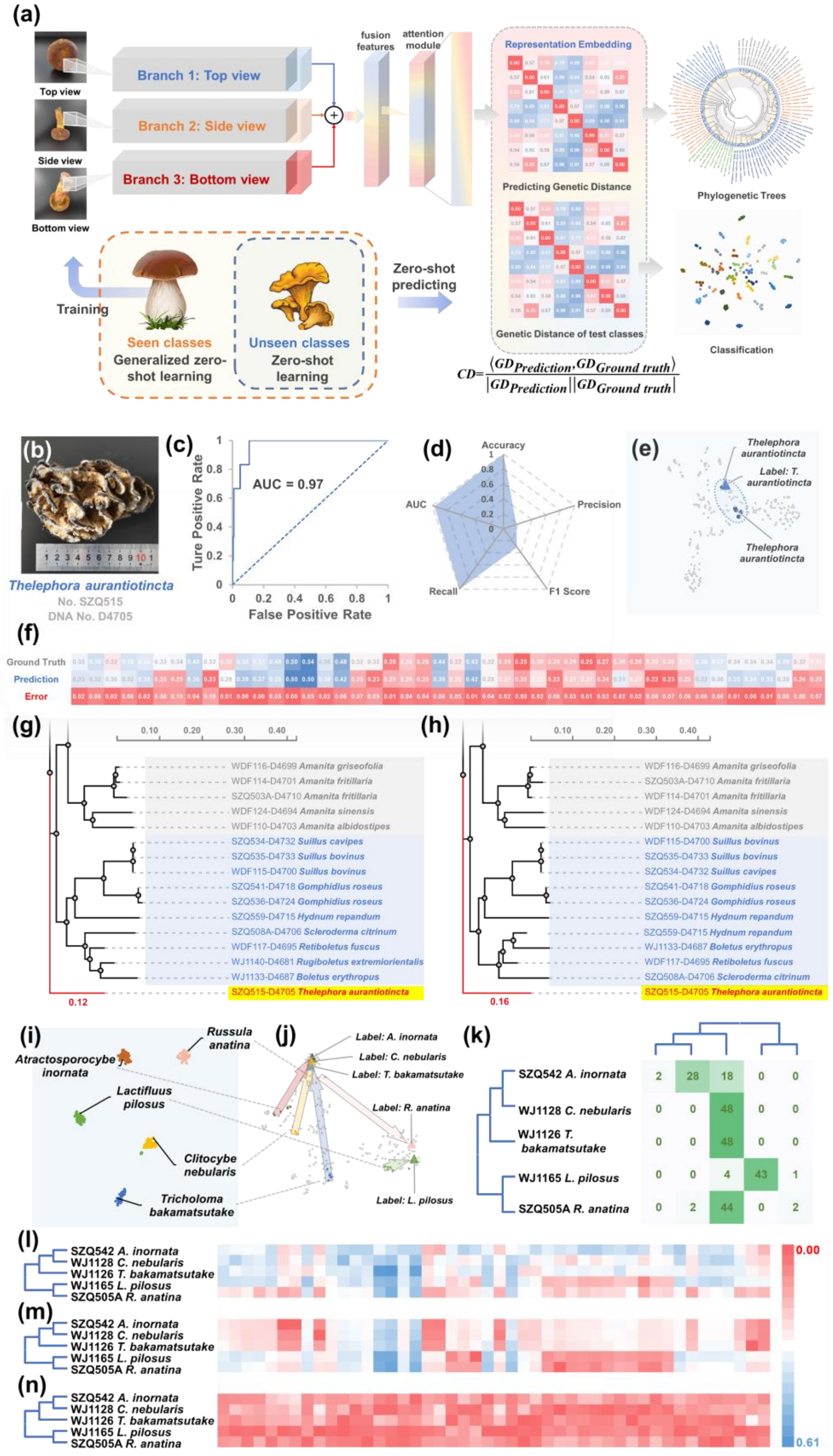

$$CD=\frac{\langle GD_{Prediction}, GD_{Ground\ truth}\rangle}{|GD_{Prediction}||GD_{Ground\ truth}|}$$

**Fig. 4 Mushroom image zero-shot recognition.**

(a) Schematic diagram of a zero-shot learning strategy based on a multi-perspective framework. The example image (b), AUC (c), rose chart (d), t-sne diagram(e), ground truth, prediction and errors of genetic distance (f), and part of N-J phylogenetic tree using predicted genetic distance (g) and ground truth of genetic distance (h) of generalized zero-shot learning for *Thelephora aurantiotincta*. The t-sne diagram of zero-shot learning(i) and generalized zero-shot learning (j) for 5 unseen species. (k) The confused matrix of zero-shot learning for 5 unseen species. The ground truth (l), prediction (m), and errors (n) of genetic distance for the prediction of 5 unseen species.

## The DNA prediction from image

In the study of DNA sequences, deep learning can use DNA sequences to predict the laboratory origin of DNA sequencing[27,28], gene-phenotype associations[29], phylogenetic relationships[30,31], and even human diseases[32]. However, there are no studies that can predict DNA sequences from bioinformatics.

In this study, based on the study of the representation space of genetic distances, we further proposed a method to map the representation space of genetic distances to DNA space, using an encoder and decoder to map genetic distances to hidden layer space and then decoding DNA by pre-trained decoder (Fig. 5a), where the total test was tested on 37 species and with DNA prediction accuracy of up to 90% (Fig. 5e). While the two strategies using softmax can barely predict DNA sequences from pictures. Specifically, our predicted DNA fragment belongs to the ITS interval, and the predicted DNA fragment consists of two parts, the A segment consisting of the second half of ITS I and the first half of 5.8S and the B segment consisting of the second half of 5.8S and the first half of ITS II, with a total length of 157 base pairs, and the cropping strategy is used in the construction of the 3000-dimensional embedding (Fig. 5d). We used *Gymnopus dryophilus*, which perfectly predicts DNA sequences, as a reference and calculated a total of 36 species with a DNA similarity of 81.01%. We used this as a baseline and found that only 3 of the remaining 36 species had an accuracy of less than 0.81 (Fig. 5b). And we counted the similarity of the first 1 to 8 base pairs and the last 125 to 157 base pairs of the non-conserved intervals as 30.74% and 54.87%, respectively (Fig. 5f). On these two non-conservative intervals, the accuracy of our prediction was 71.85% and 71.96% (Fig. 5g-h). On the conserved intervals, our accuracy of DNA prediction was 95.97%, which was higher than the baseline of 91.91%. It shows that our method did not overfit to a specific sequence and did not lose its predictive power in non-conserved regions (the expected value of the accuracy of randomly generated 4 base pairs is 25%) and predicted the DNA sequence well.

Zero-shot learning was tested on two *Theleora aurantiotincta*, *Amanita vagunata* unseen species with DNA accuracy at 87.47% and 83.43%, respectively (Fig. 5i and Supplementary Table 21). However, zero-sample DNA prediction is less effective in non-conserved regions, probably due to the poor generalization of the decoder on only 37 DNA sequences. Therefore, using larger datasets and training decoders with higher accuracy to predict genetic distances is a way to improve the effectiveness of predicting DNA sequences from images.

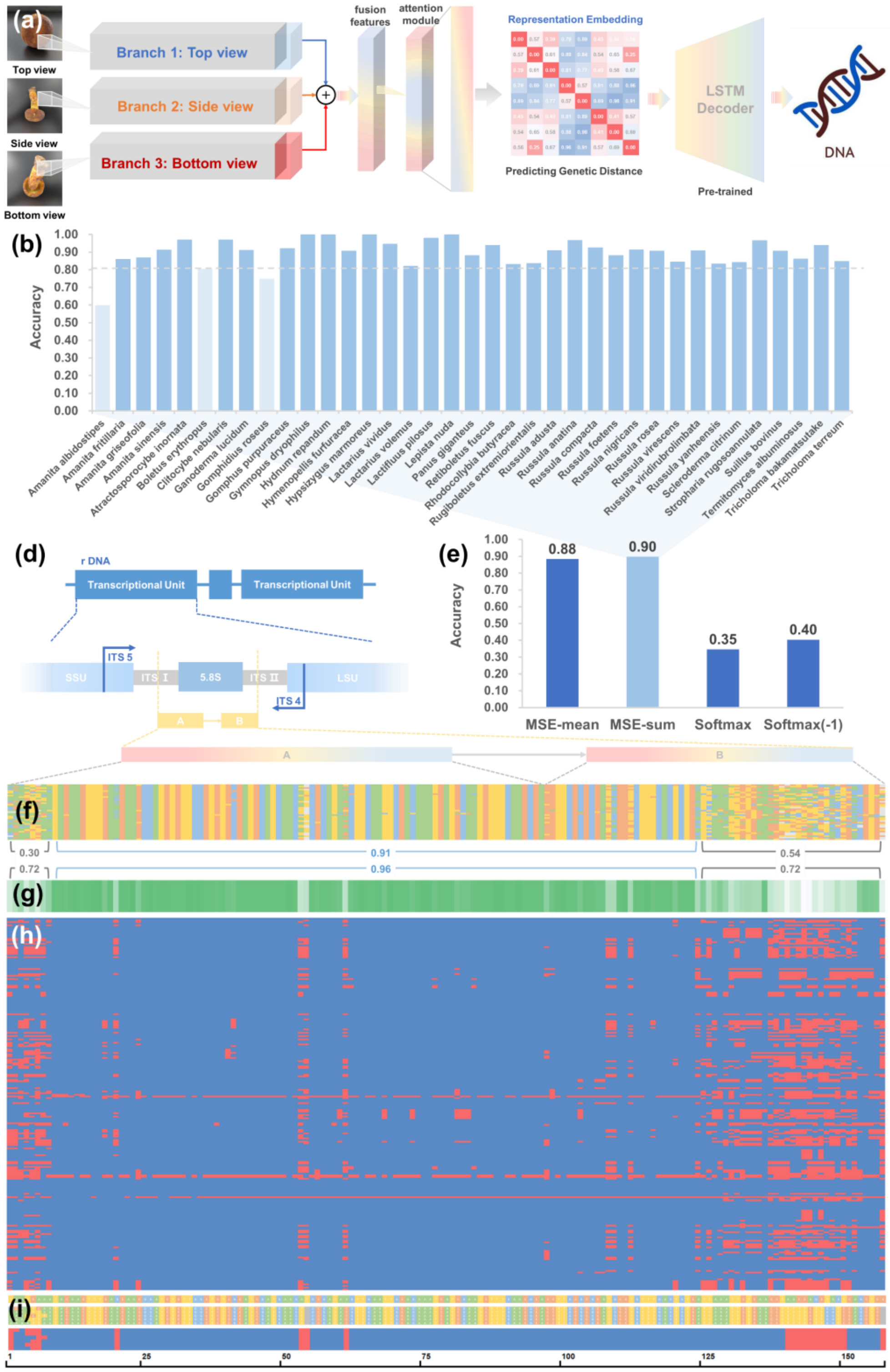

**Fig. 5 DNA prediction based on mushroom images.**

(a) Schematic diagram of the process of inferring DNA from mushroom images. (b) The accuracy of DNA prediction for 37 different species using ResNet-50 backbone inducing MSE-sum active function, the gray dashed line represents the baseline accuracy of 0.81, and the accuracy of DNA for different active functions (e). The schematic diagram of the predicted ITS region of DNA (d), heatmap (f) of 37 species' DNA ( green is A, yellow is T, blue is G, and red is C). The heatmap of total DNA accuracy (g) and per-base accuracy (h) (bule is right, red is false) of 222 DNAs of 37 species predicted by using the ResNet-50 backbone with MSE-sum active function. (i) The heatmap of label DNA, predicted DNA, and per-base accuracy of zero-shot learning for *Thelephora aurantiotincta.*

## Discussion

In this work, we propose a multi-view image recognition framework mediated by genetic distances to achieve predictions of taxonomy and evolution on mushrooms.

We demonstrate that images from different viewpoints do not differ significantly in recognition accuracy. However, our proposed multi-perspective image recognition framework significantly improves the recognition ability by fusing features from different views, which is not only useful for the fine-grained recognition problem of mushrooms, but also can be further extended to the whole field of fine-grained image recognition.

The striking part in this study is our incorporation of genetic distance as a mediator in the image recognition framework. Genetic distance adds information with biological and evolutionary semantics to the biological image recognition process. Not only mushroom images, but any biologically related image recognition can be associated with genetic distance, such as fine-grained images of microorganisms, birds, insects, plants, etc., medical images[33] such as X-ray[34], CT[35], microscopic pathology images[36,37], human face images[38], biometric images, etc.

The introduction of genetic distance also brings inspiration for the interpretability of deep learning models. Compared to the traditional embedding of text and words which are parameters extracted from classifiers, genetic distance embedding itself has biological significance. With the help of genetic distance, we found that softmax overestimates the distance of samples in the feature space in traditional pattern recognition tasks, and adding genetic distance to softmax still enables the model to maintain a semantic topology. Therefore, we envision that the genetic distances of biological species pairs can be extended to the category of arbitrary category pairs, forming category distances for introducing advanced semantics into pattern recognition tasks. Moreover, the distances generated based on category labels do not require the involvement of datasets, making the method simple and convenient and allowing the introduction of a large amount of semantic information.

The excellent accuracy and generalizability of our model for predicting genetic distances of images gives our method the ability to predict unknown kinds of categories and evolutionary information. This will bring new inspiration and reference to future taxonomy and phylogenetics. Based on our method, it is possible to design fast and convenient identification algorithms to count biodiversity in the field in the future, and also to assist in the discovery of new species.

By mapping macroscopic features to genetic distances, we break the barrier between morphology and molecular phylogenetics, and bring inspiration to resolve the contradiction between morphological features and molecular information in phylogenetics. We also demonstrate that macroscopic morphological characters can be well identified to the genus level, but it is still difficult to identify species at the species level, and the introduction of microscopic morphological characters is needed to improve the accuracy in the future.

Excitingly, we have predicted DNA sequences from images for the first time and showed excellent predictive power. We envision that in the future, further development based on our method can rapidly acquire preliminary DNA sequences from biological images and combine them with other deep learning algorithms for fast and high-throughput prediction of protein structures[18,39], protein allosteric sites[40], disease[41], etc. If successful, the methods of image recognition can be used to assist sequencing means and greatly reduce the cost of large-scale sequencing. Combined with modern powerful prediction capabilities for proteins, this ability to predict DNA sequences from images will provide a powerful tool for biological, medical, industrial, agricultural and environmental fields, and revolutionize the future biological landscape.

## Methods

### ***Multi-perspective mushroom dataset***

In this study, a total of 246 mushroom specimens of 49 taxa, 40 species, and 24 genera were collected in Guizhou between September and December 2021, and images were collected for both sides, three axes, and three views of the specimens. A total of 4428 mushroom images were collected, and the details of images for 40 species are shown in Supplementary Tables 1. Morphological identification of mushrooms was performed for all of them. The DNA of the ITS interval of each specimen was also sequenced, and the molecular identification of the mushrooms was performed based on the sequencing results. The identification results and collection information are shown in Supplementary Table 2. The mushroom specimens were preserved in the mycological herbarium of the Institute of Microbiology, Guizhou Provincial Academy of Sciences.

The reference sequences corresponding to the multi-view mushroom dataset are shown in Supplementary Table 3.

***Coding experiment environment***

In the experiments of this project, the main algorithms such as data enhancement and CNN were performed using Anaconda3 (Python 3.6), PyTorch GPU library, and OpenCV-python3 library, except for some image pre-processing performed by Photoshop tools. CNN training and testing were accelerated by GPU, and the experimental hardware configuration included Intel® Xeon(R) E5-2620 CPU (2.10 GHz), 64 GB RAM, and NVIDIA GeForce RTX 2080 (CUDA 10.2) graphics card for model training and testing.

***Three-stage training for multi-perspective framework***

In this task, we propose three phases of multi-perspective model training. In the first stage, we initialize the model parameters and load the parameters and weights pre-trained from ImageNet by transfer learning. In the second stage, separate branch models are trained on input image data from different viewpoints, the classification layers of the branch models are replaced with those suitable for our mushroom dataset, and the models are trained on our dataset to fine-tune the parameters and weights and to retrain the classification layers. In the third stage, after all three branches of the model are trained, we discard the classification layer, fuse the features of the three branches using concat, and then classify the fused features by a classifier at the end. In the third training step, we freeze all parameters of the branches and only fine-tune the classifier, which consists of an SE module, an ECANet module, and a linear layer.

The first stage uses migration learning techniques. By training on the ImageNet dataset, parameters and weights are loaded into the model to initialize it.

In the second stage, the branching models are trained separately in order to fit our dataset, we replace the classification layer, use the Adam optimizer to train and update the parameters and weights, and set the learning rate to 0.0001.

In the third stage, we fused the branching models and added the attention module, froze all the layers in the model except the last convolutional layer, the attention module (SE and ECANet) and the softmax layer, trained the attention module with our dataset, updated the weights with the Adam optimizer and set the learning rate to 0.0001.

***Comparison of different backbone models in multi-perspective framework***

For branching models, a branch only needs to focus on the information of one angle of the mushroom, for different backbone models, the focus of different models may not be the same, here, this project selected several classical models in computer vision, VGG, Squeezenet, Exception, Densenet 121, Resnet50, ResNext50, ShuffleNet-V2, MobileNet-V2, MobileNet-V3, and EfficientNet-B0, to investigate the performance effects of different backbone models under Multi-perspective structure.

In order to explore the effect of the number of layers of branching networks on the classifier, here, 18, 34, 50, 101, and 152 layers of ResNet are selected as backbone for this project.

***Study of fine-grained image recognition granularity based on multi-perspective framework***

The division of different levels in taxonomy may have an impact on the accuracy of classification tasks, especially for fine-grained classification problems, where the division of classification levels is the granularity level of fine-grained classification. In this study, samples at the genus level of hierarchy (Supplementary Table 4) were selected from the multi-perspective mushroom image dataset for the classification task. nine species under the genus *Russula* (Supplementary Table 5) were studied for intra-genus interspecific classification.

***Calculation of genetic distance for multi-perspective dataset***

The multi-view mushroom image dataset corresponding to ITS sequencing results are shown in Supplementary Table 6, while the corresponding reference sequences were downloaded from NCBI as shown in Supplementary Table 7. In this study, the sequenced sequences were aligned with the reference sequences simultaneously. The sequences were aligned using Clustal W of MEGA11, with a base pair alignment Gap open penalty of 15 and a Gap extension penalty of 6.66, and a multiple sequence alignment Gap open penalty of 15 and a Gap extension penalty of 6.66. The DNA matrix weight was IUB and the exchange weight was 0.5. Because of the large number of species, the aligned sequences were trimmed using PhyloSuite's The aligned sequences were trimmed using Gblocks with a maximum length of 10 in the A region and a maximum neighboring non-conserved site of 8. The aligned trimmed sequences were genetically distanced using Pairwise Disantance, and uncertainty calculations were performed using 100 Bootstrap samples. The replacement model was calculated using Nucleotide, using the Maximum Composite Likelihode method.

The above genetic distances were calculated based on a two-by-two comparison of the genetic distances of species in the dataset species. This study further introduced species outside the dataset to participate in genetic distance generation, where the species composition for genetic distance calculation of length 100 is shown in Supplementary Table 9.

The DNA sequences with a genetic distance of 3000 were obtained from 3000 ITS sequences of Agaricomycetes under the Fungal Internal Transcribed Spacer RNA (ITS) RefSeq Targeted Loci Project of NBCI. Sequences were aligned using the EBI (https://www.ebi.ac.uk/) online alignment Cluster Omega tool.

Genetic distances of length 20 and 10 were obtained by selection in Supplementary Table 6.

***Phylogenetic analysis by DNA***

The aligned sequences were used to construct a phylogenetic tree using the Maximum Likelihood Tree method, the phylogenetic tests were sampled 100 times using the Bootstrap method, and the replacement model was calculated using the Tamura-Nei model using Nucleotide replacement.

***Image distance generation based on genetic distance***

This study species uses genetic distance as the Representation space and uses the genetic distance between two mushroom species within the dataset as the Embedding of genetic distance representation space. In this study, image features are extracted using convolutional neural networks, and image features are mapped to the feature space of the fully connected layer (fc). The feature space and the representation space are connected by softmax or MSE functions, i.e., Embedding of genetic distance. by learning on the training set, the model gains the ability to generate genetic distance by image features, i.e., image distance with biogenetic information is obtained by feature extraction.

***Applying Genetic Distance for Mushroom Identification***

In this study, the Embedding of the genetic distance predicted by the model and the Ground truth of the genetic distance (including cosine distance, Euclidean distance, etc.) are subjected to distance operation, and the obtained distance is Label Embedding, which can be mapped to Label space (Label Sapce) by Label Embedding to complete Mushroom image classification.

***Zero-sample learning based on genetic distance***

In this study, based on the aforementioned genetic distances predicted by image features, a wide range of features of mushroom species can be learned, and features can be mapped to a genetic distance representation space containing species other than the training species, thus allowing zero-sample learning.

The Embedding generated by image feature space mapping can be compared with the genetic distances of species within the non-training set to calculate cosine distances, Euclidean distances, etc., so that a zero-sample Label Embedding can be generated directly and mapped to the zero-sample label space.

***Prediction of DNA sequences from mushroom images***

In this study, the genetic distance generated by the model is used to predict the DNA sequences of species. First, the images are transformed into 100-dimensional genetic distances using a multi-branch model, and the 100-dimensional genetic distances are processed by two layers of LSTM. The results obtained from the encoder are fed into the decoder (LSTM-based, 2-layer), which is given a start flag and cycles through the output DNA sequences until an end flag is encountered.

**Acknowledgments**

This work was supported by the National Undergraduate Training Programs for Innovation and Entrepreneurship (202110022063), the Projects of Science and Technology of Programs of Guizhou Province ([2019]-4007, [2018-4002]), and the Fifth Batch of "Thousands of Innovative and Entrepreneurial Talents" in Guizhou Province.

**Data availability**

Our multi-perspective mushroom datasets are available from https://www.kaggle.com/datasets/jiewenxiao/multiinput-mushroom-classification.

**Code availability**

Our codes are available from https://github.com/set-path/multi-input-mushroom-recognition.

**Contributions**

The work was conceptualized by J. X. and J.-X. W. In details, J. X., W. L., design and built the muti-perspective recognition framework with supervision and advice from J.-X. W. and Y. Y. The muti-perspective mushroom dataset was constructed by J. X., J. W., and Y. Y., and identified by M. Z. J. X. proposed the genetic information meditating methods. W. L. and J. W. developed the approach for DNA prediction from image. W. L. wrote the software. J. X. designed the experiments, analyzed and visualized the data, wrote the original draft. J.-X. W., J. X., W. L., M. Z., and J. W. reviewed and edited the manuscript. J.-X. W. and Y. Y. supervised the project.